\documentclass[10pt,twocolumn,letterpaper]{article}

\usepackage{cvpr}              

\usepackage{algorithm}
\usepackage{algpseudocodex}
\usepackage{amsmath}
\usepackage{amssymb}
\usepackage{multirow}
\usepackage{booktabs}
\usepackage{makecell}
\usepackage{subcaption}
\usepackage{fancybox}
\usepackage{fancyvrb}
\usepackage{bm}
\usepackage{multirow}
\usepackage{booktabs}
\usepackage{bbding}
\usepackage{makecell}
\usepackage{subcaption}
\usepackage{fancybox}
\usepackage{fancyvrb}

\definecolor{cvprblue}{rgb}{0.21,0.49,0.74}
\usepackage[pagebackref,breaklinks,colorlinks,citecolor=cvprblue]{hyperref}

\def\ours{RealVVT}

\title{RealVVT: Towards Photorealistic Video Virtual Try-on via \\ Spatio-Temporal Consistency}

\author{
  Siqi Li$^{1}$ \quad Zhengkai Jiang$^{2}$ \quad Jiawei Zhou$^{3}$ \quad Zhihong Liu$^{4}$ \quad Xiaowei Chi$^{2}$ \quad Haoqian Wang$^{3\dag}$\\
  $^1$Intellifusion \;\; $^2$HKUST \;\; 
   $^3$THU \;\;
  $^4$FDU \\
  \texttt{\small tristanafourseven@gmail.com} \quad
  \texttt{\small wanghaoqian@tsinghua.edu.cn} 
}

\begin{document}

\twocolumn[{%
\renewcommand\twocolumn[1][]{#1}%
\maketitle
\begin{center}
    \centering
    \captionsetup{type=figure}
    \includegraphics[width=0.82\textwidth]{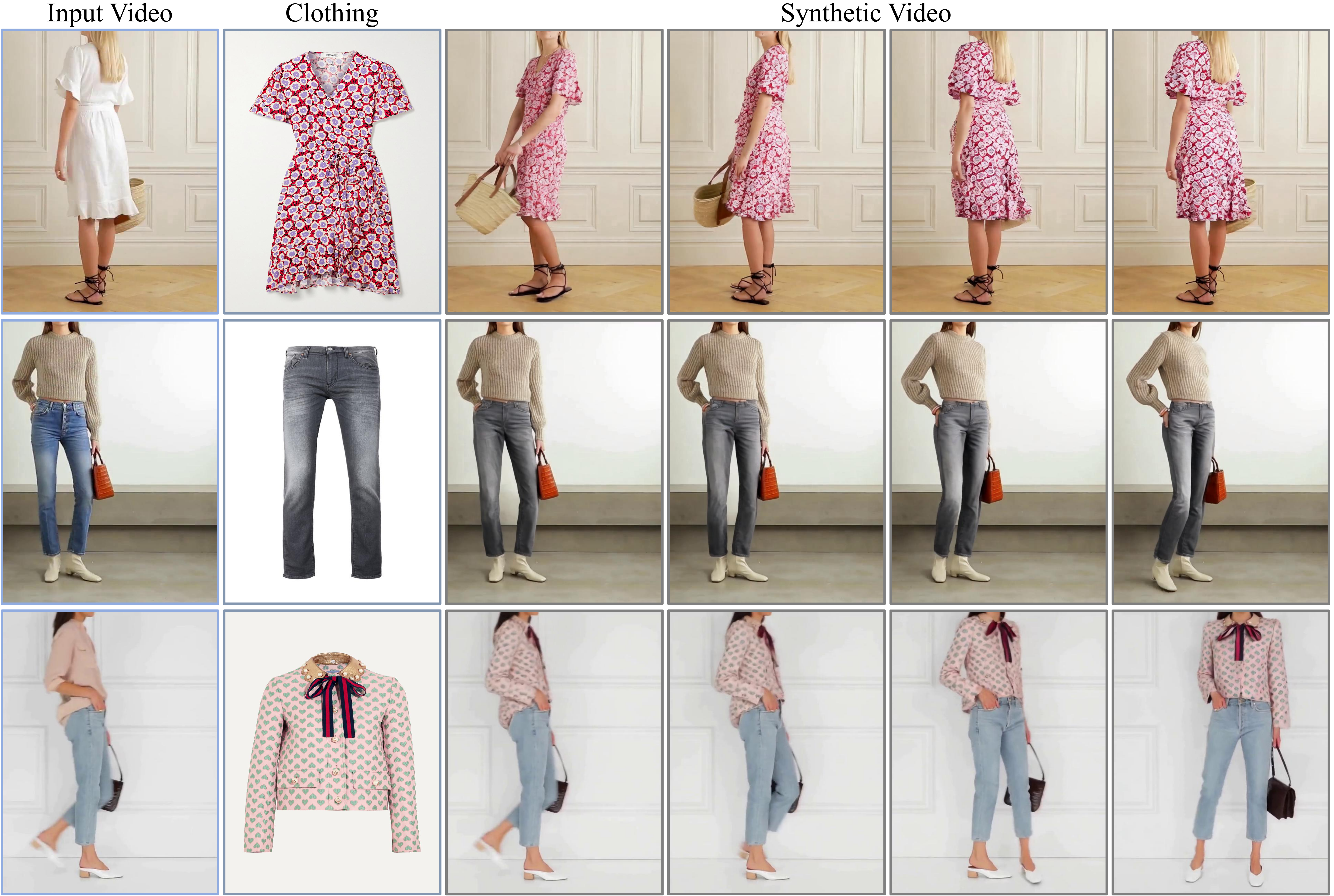}
    \captionof{figure}{RealVVT is a novel framework that takes as input a video of a human performing arbitrary motions from any viewpoint, along with a garment (e.g., upper body, lower body, or dress) to be virtually worn. The system seamlessly integrates the garment into the person's “OOTD”(Outfit Of The Day) and evaluates its aesthetic compatibility and fit through dynamic video results. This figure showcases a subset of generated results, demonstrating RealVVT's ability to maintain the characteristics and details of the target garment while ensuring consistency with the subject's motion. \vspace{1em}
    }
    \label{fig:teaser}
\vspace{-1mm}
\end{center}%
}]
{\let\thefootnote\relax\footnotetext{{$^{\dag}$Corresponding authors.}}}

\begin{abstract}
Virtual try-on has emerged as a pivotal task at the intersection of computer vision and fashion, aimed at digitally simulating how clothing items fit on the human body. Despite notable progress in single-image virtual try-on (VTO), current methodologies often struggle to preserve a consistent and authentic appearance of clothing across extended video sequences. This challenge arises from the complexities of capturing dynamic human pose and maintaining target clothing characteristics. We leverage pre-existing video foundation models to introduce \textit{RealVVT}, a photoRealistic Video Virtual Try-on framework tailored to bolster stability and realism within dynamic video contexts. Our methodology encompasses a Clothing \& Temporal Consistency strategy, an Agnostic-guided Attention Focus Loss mechanism to ensure spatial consistency, and a Pose-guided Long Video VTO technique adept at handling extended video sequences. Extensive experiments across various datasets confirms that our approach outperforms existing state-of-the-art models in both single-image and video VTO tasks, offering a viable solution for practical applications within the realms of fashion e-commerce and virtual fitting environments.
\end{abstract}
    
\section{Introduction}
\label{sec:intro}

With significant advancements in image-based virtual try-on (VTO) technology, there has been a growing demand for video virtual try-on (VVT), driven by the need to capture and display the dynamic appearance of clothing on individuals across video sequences. VVT not only preserves fine-grained garment details but also ensures that clothing aligns naturally with the wearer’s motions and body shapes, providing users with an immersive experience to visualize how desired clothing fits and moves from various perspectives. This innovation has attracted considerable attention for two key reasons: first, its practical applications in the fashion industry and entertainment, and second, its potential to inspire new directions in video editing tasks based on image prompts. These factors have collectively accelerated progress in this rapidly evolving field.

The task of VVT presents significant challenges, extending beyond static operations like mapping garments to predefined masks. Unlike static images, VVT must handle dynamic human poses and varying viewpoints, complicating the accurate fitting of clothing to the target individual. Movement and perspective changes can distort the garment's appearance, making it difficult to preserve its shape, style, and texture. Additionally, ensuring spatial and temporal consistency throughout the video sequence is critical for successful VVT. Previous approaches~\cite{jiang2022clothformer, dong2019fw, zhong2021mv} have addressed these challenges using optical flow estimation and completion techniques, where garments are warped using optical flow and misalignments are corrected via generative mechanisms like GANs~\cite{dong2019fw} or Transformers~\cite{jiang2022clothformer}. Recently, diffusion-based methods~\cite{wild_vid_fit_he2024wildvidfit, vivid_fang2024vivid, xu2024tunnel, zheng2024viton} have emerged, adapting text-to-image techniques to the video domain. These methods~\cite{vivid_fang2024vivid, xu2024tunnel, wild_vid_fit_he2024wildvidfit}, which incorporate temporal modules or consistency constraints, show promise for VVT. Some approaches~\cite{zheng2024viton} have even adapted text-to-video diffusion models directly, demonstrating the potential to model garment transformations according to the target individual's poses and motions.
However, VVT still faces three primary challenges:
Spatial Inconsistency: Garments tend to adhere to the target mask's shape and color, failing to preserve the original garment's shape, style, and texture.
 \textbf{Spatial Inconsistency}: Garments tend to adhere to the target mask's shape and color, failing to preserve the original garment's shape, style, and texture. 
\textbf{Temporal Inconsistency}: Maintaining consistent clothing appearance during movement remains difficult, often resulting in flickering or unstable garments.\textbf{Long Video Inaccuracy}: Frame-by-frame generation accumulates errors over time, especially with complex body movements and occlusions, leading to unexpected outcomes and cumulative inconsistencies.

 To address these challenges, we propose RealVVT (Realistic Video Virtual Try-on), a novel framework that leverages the strengths of diffusion models, which have recently shown remarkable success in image and video generation tasks. Built on Stable Video Diffusion (SVD)~\cite{blattmann2023stable}, a state-of-the-art image-to-video architecture, RealVVT incorporates several key innovations to ensure high-quality, temporally coherent virtual try-on results.

To enhance video generation quality, we focus on improving spatial and temporal consistency. For spatial consistency, we introduce the \textbf{Agnostic Mask-Guided Attention Loss}, which ensures accurate intra-frame garment fitting by directing the model to prioritize wearable regions while reducing attention to non-wearable areas. This preserves the garment’s shape, style, and texture, while appropriately filling mask regions to maintain spatial alignment and authenticity across the sequence.

For \textbf{temporal consistency}, we propose the \textbf{Clothing \& Temporal Consistency Attention} mechanism, which leverages interactive information between two U-Nets~\cite{hu2024animate} to integrate reference and temporal data. This ensures stable clothing alignment with the wearer’s body, even during pose changes or camera shifts, significantly reducing flickering and misalignment.

To address long video inaccuracies, we introduce the \textbf{Pose-guided Long VVT} strategy, which uses pose inputs to estimate motion and viewpoint changes. By iteratively generating long videos through keyframe replacement, this strategy preserves motion realism and clothing coherence throughout the sequence.

Experiments on multiple high-quality image and video datasets demonstrate the superior performance of our approach in both short- and long-video virtual try-on tasks. \cref{fig:teaser} showcases our generated results.

In summary, the contributions of this paper are threefold:
\begin{itemize}
    \item Agnostic Mask-Guided Attention Loss, which focuses on garment-wearing areas while minimizing attention to non-wearable regions, ensuring accurate spatial alignment.
    \item Clothing \& Temporal Consistency Attention Mechanism, which integrates reference and temporal information across frames to reduce flickering and misalignment, enhancing temporal coherence.
    \item Pose-guided Long VVT strategy for long video sequences, preserving motion realism and garment coherence by iteratively video generation.
\end{itemize}

 \section{Related Work}
\label{sec:related}

\begin{figure*}[t]
\centering
\includegraphics[width=0.96\textwidth]{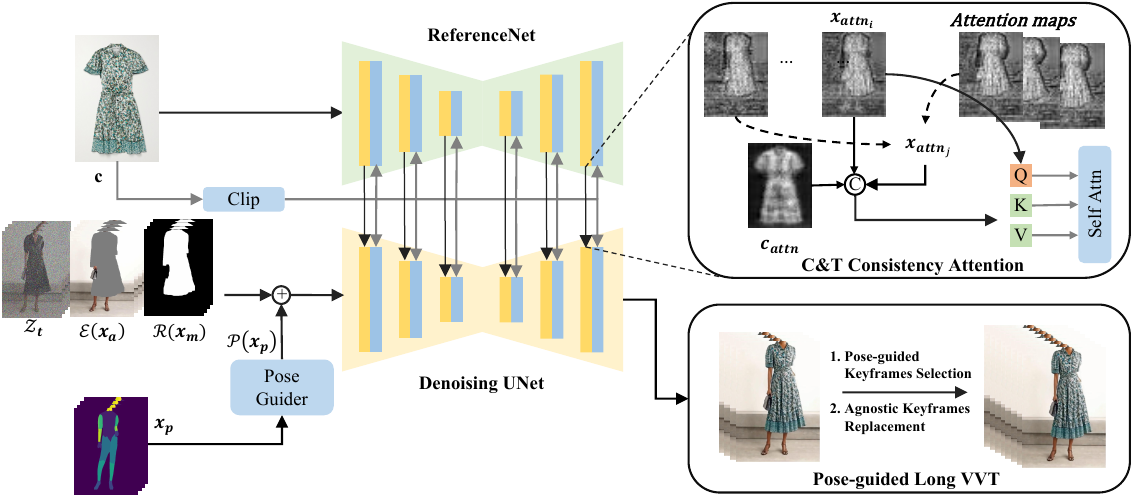}

\caption{\textbf{An overview of RealVVT}. A Reference Net and CLIP Encoder extract target garment features, while the input video is processed by Denoising UNet. The figure omits the VAE encoder and decoder for clarity. The right side illustrates the mechanisms of our proposed Clothing \& Temporal Consistency Attention and Pose-guided Long VVT components. }
\label{fig:main}
\vspace{-0.3cm}
\end{figure*}

\subsection{Video Virtual Try-On}
Video try-on aims to transfer a garment onto a target individual while preserving the garment’s shape and visual details over time as the person moves or changes perspective. Existing approaches to video virtual try-on can be broadly categorized into GAN-based~\cite{dong2019fw, zhong2021mv, kuppa2021shineon, jiang2022clothformer} and diffusion-based methods~\cite{xu2024tunnel}. GAN-based methods typically depend on optical flow to warp the garment~\cite{dosovitskiy2015flownet} and employ a GAN generator to blend the warped garment with the person. GAN-based models are sensitive to misalignment between the garment and the person due to inaccurate flow estimations and often lag behind diffusion-based models in generation quality due to the latter’s use of large-scale pretrained weights. The era of diffusion has arrived, Tunnel Try-on~\cite{xu2024tunnel} employs a UNet-based model for video try-on, enabling it to handle camera movements and accurately preserve clothing textures. ViViD~\cite{vivid_fang2024vivid} introduced a high-resolution dataset (832 × 624) for video try-on, addressing the limitation of prior datasets like VVT~\cite{bai2022single}, which offered only low-resolution samples. VITON-DiT~\cite{zheng2024viton} generates try-on sequences in-the-wild settings by using the DiT structure~\cite{li2022dit}. However, its text-to-video architecture is redundancy and inefficient. Meanwhile, WildVidFit~\cite{wild_vid_fit_he2024wildvidfit} employs a two-stage, image-generation-based framework for try-on, trained in two separate stages and lacking the abilities of temporal coherence and preserving fine details. Building upon these prior approaches, we introduce RealVVT, a one-stage training framework for video try-on that achieves high-quality synthesis with superior spatio-temporal consistency, and maintaining efficiency.

\subsection{Video Generation via Diffusion Models}
With continued advancements in diffusion-based image synthesis techniques~\cite{dhariwal2021diffusion, kang2024distilling, kumari2023multi}, numerous frameworks have been developed to extend diffusion models for video synthesis. Some approaches train video diffusion models from scratch by incorporating temporal layers~\cite{he2022latent, zhang2024trip}, while a more prevalent strategy involves adapting pretrained image diffusion models for video by adding temporal layers and fine-tuning them specifically for video generation tasks~\cite{guo2023animatediff, ma2024latte, yin2023nuwa}. However, these approaches still face challenges struggle with maintaining fine-grained texture consistency and temporal coherence across frames, especially when complex garment details need to be preserved throughout varying poses and movements in a video sequence. The DiT structures~\cite{liu2024sora, jiang2024dive, chen2024gentron} can effectively capture spatio-temporal dependencies, their computational demands increase significantly with higher resolutions and longer sequences, posing challenges for high-fidelity video try-on applications. SVD~\cite{blattmann2023stable} exemplifies this approach: it builds upon a latent image diffusion model~\cite{rombach2022high} and is adapted to video synthesis with additional temporal components, including 3D convolutions and temporal attention layers. SVD is well-suited for maintaining high levels of spatial and temporal coherence, as it leverages pretrained image diffusion capabilities while incorporating temporal modeling to ensure frame-to-frame consistency. Building on the large-scale pretrained SVD model, we introduce a new method that achieves significantly enhanced spatial and temporal consistency compared to prior models.

\section{Proposed Approach}
\label{sec:method}

We first review some foundational concepts of video diffusion models in \cref{sec:preliminary}. Following this, \cref{sec:overview} offers a detailed overview of the overall network architecture of our \ours model. In \cref{sec:agnostic_attention}, we introduce an Agnostic-guided Attention Focus Loss, which can improve spatial consistency. Subsequently, in\cref{sec:temporalconsistency}, we describe our method to achieve temporal consistency. Finally, we prensent a  pose-guided strategy in \cref{sec:longvvt} for long video virtual try-on generation.
\subsection{Preliminary}
\label{sec:preliminary}
Our primary UNet backbone leverages Stable Video Diffusion~\cite{blattmann2023stable} model to jointly train videos and images in a unified framework, utilizing the EDM~\cite{karras2022elucidating} diffusion model with Euler-step sampling strategy.

\noindent\textbf{Video Diffusion Model}. Stable Video Diffusion (SVD) was initially developed for the purpose of video generation, leveraging a single image as the initial frame. This model consists of a Variational Autoencoder (VAE)~\cite{kingma2013auto} and a UNet architecture that incorporates spatio-temporal blocks. The VAE encoder transforms input video frames into a lower-dimensional latent space, while the decoder reconstructs these latent representations back into the frame space. To mitigate temporal inconsistencies and reduce flickering artifacts, temporal layers are integrated within the VAE. In the latent space, a conditional spatio-temporal U-Net is employed for denoising, utilizing both spatial and temporal information through 3D convolutional layers. This architecture effectively integrates conditional inputs to enhance the denoising process.

\noindent\textbf{EDM.}  
In Stable Video Diffusion (SVD), the denoiser \( D_{\theta} \) receives the clean image from the outputs of the UNet \( U_{\theta} \):
\begin{equation}
\resizebox{0.9\linewidth}{!}{$
   D_{\theta}(x;\sigma,c) = c_{skip}(\sigma)\cdot x + c_{out}(\sigma)\cdot U_{\theta}(c_{in}(\sigma)\cdot x; c_{noise}(\sigma), c),
$}
\end{equation}
where \( \sigma \) denotes the noise level of the distribution, while \( c_{skip}(\sigma) \), \( c_{out}(\sigma) \), \( c_{in}(\sigma) \), and \( c_{noise}(\sigma) \) are EDM preconditioning parameters that depend on the noise level. The variable \( c \) represents the conditional input (e.g., the first frame in SVD and cloth information in our approach). As a training loss, SVD employs a continuous-time diffusion framework, EDM, in conjunction with the Denoising Score-Matching (DSM) loss function to train the denoiser \( D_{\theta} \):
\begin{equation}
\resizebox{0.9\linewidth}{!}{$
\mathbb{E}_{(x_0, c) \sim p_{\text{data}}, (\sigma, n) \sim p(\sigma, n)} \left[ \lambda_\sigma \left\| D_{\theta}(x_0 + n; \sigma, c) - x_0 \right\|_2^2 \right],
$}
\end{equation}
here, \( p(\sigma, n) \) represents the distribution of the noise level \( \sigma \) and normal noise \( n \), while \( \lambda_\sigma \) denotes the loss weights across different noise levels.

\subsection{Overview}
\label{sec:overview}
Given a video sequence $x\in\mathbb{R}^{N \times H \times W \times 3}$ of a person , where $N$ signifies the frame count during the training phase. 
The segmentation mask~\cite{xu2024ootdiffusion} of the garment to be removed is denoted $x_m \in\mathbb{R}^{N \times H \times W \times 3}$. 
The Clothing-Agnostic Person Representation ~\cite{choi2021viton} $x_a \in\mathbb{R}^{N \times H \times W \times 3}$ is obtained through a pixel-wise operation, $x_a = x \circledast x_m$, referred to as \textbf{agnostic video}, which is designed to eliminate the garment intended for replacement within $x$. Given another garment $c\in\mathbb{R}^{H \times W \times 3}$, which is intended to be worn by the person in $x$, and evaluated for compatibility in his or her "OOTD". $c$ belongs to the same category of clothing as $x_m$, but typically differs in shapes and textures. 
We frame the video virtual try-on task as an exemplar-based video inpainting problem. The goal is to fill the agnostic mask region $x_m$ in the agnostic video $x_a$ with the target garment $c$, while leveraging the unmasked regions of $x_a$ to provide complementary information about the individual, such as exposed skin or other visible clothing. 

As illustrated in \cref{fig:main}, the framework is built upon a dual U-Net structure, consisting of \textbf{Denoising UNet} and \textbf{ReferenceNet}, which has been proven effective in virtual try-on methods~\cite{xu2024tunnel, zheng2024viton, vivid_fang2024vivid, wild_vid_fit_he2024wildvidfit, wang2024gpd}. The ReferenceNet is employed to encode the fine-grained features of $c$, initialized using Stable Diffusion (SD). Concurrently, the Denoising UNet is primarily responsible for denoising the video sequence of the target person, initialized using Stable Video Diffusion (SVD).
For the Denoising UNet's input, we combine three components: (1) the noisy frames ($Z_t$, 4 channels); (2) the latent agnostic video frames ($\mathcal{E}(x_a)$, 4 channels); and (3) the resized cloth agnostic masks ($\mathcal{R}(x_m)$, 1 channel). To unify the input channels, we modify the initial convolutional layer of the U-Net to accept 9 channels (\eg, $4+4+1=9$). Additionally, the model incorporates dense pose information($\mathcal{P}(x_p)$) from the pose guider~\cite{hu2024animate}, which helps that the denoising process preserves the individual's motion and posture. And the model is conditioned on the reference cloth image figure provided by the ReferenceNet and CLIP~\cite{radford2021learning}.

\subsection{Agnostic Mask-Guided Attention for Clothing Consistency}
\label{sec:agnostic_attention}
In the attention mechanism of SVD, the attention probability scores, denoted as $S$, represent the distribution of attention weights across different regions, where higher values indicate areas of greater focus. To facilitate the replacement of the original clothing with the target garment, we aim to ensure that the regions corresponding to the agnostic mask $x_m$ receive heightened attention for the target garment. To achieve this, we propose a novel loss function designed to enhance attention efficacy specifically within the agnostic mask region. The initial formulation is as follows:

\begin{equation}
\mathcal{L}_{\text{agn-init}} = \sum_{i \in N} \sum_{a \in A} (1 - S_i^a)^2 + \lambda_N \sum_{i \in N} \sum_{a \in \bar{A}} \|S_i^a\|^2 ,
\end{equation}
where $N$ denotes the length of the video sequence, $A$ represents the highlight region defined by the agnostic mask $x_a$, and $\bar{A}$ denotes its complement. Here, $S_i^a$ corresponds to the attention probability for the target garment at location $a$ in frame $i$. This loss function encourages higher attention probabilities within the agnostic mask region $A$ for the target garment, while simultaneously suppressing attention in non-mask regions $\bar{A}$. The parameter $\lambda_n$ ontrols the trade-off between positive (mask region) and negative (non-mask region) attention contributions, where $n$ indicates its application to the negative component.
However, our goal extends beyond simply filling the mask region with the target garment; it involves replacing clothing A with clothing B, which may differ in shape, style, or coverage. In practice, the agnostic mask rarely aligns perfectly with the target garment, especially when the replacement involves significant style changes (e.g., pants to shorts or short-sleeve to long coats). In such cases, accurately positioning clothing B within the mask region is critical. Furthermore, the model must infer and fill the areas outside B but within $x_m$ with contextual details, such as skin tone or limb shape, to realistically reconstruct occluded body parts.

To address these challenges, we refine the loss function to prioritize attention in the most relevant regions, rather than uniformly across the entire mask. This guides the model to focus on areas with high attention probability. The revised loss function is defined as:
\begin{equation}
\mathcal{L}_{\text{agn}} = \sum_{i \in N} (1 - \max_{a \in A} S_i^a)^2 + \lambda_N \sum_{i \in N} \sum_{a \in \bar{A}} \|S_i^a\|^2 .
\label{eq:loss4}
\end{equation}

This enhances the model's ability to infer context beyond the agnostic mask boundaries, better preserving the original clothing characteristics while ensuring smooth transitions between the generated regions and the surrounding areas. Finally, we fine-tune RealVVT by incorporating $\mathcal{L}_{\text{agn}}$ to $\mathbb{E}_{(x_0, c)}$ :
\begin{equation}
\mathcal{L}_{\text{modified}} = \mathbb{E}_{(x_0, c)} + \lambda_{\text{agn}} \mathcal{L}_{\text{agn}} . 
\label{eq:loss5}
\end{equation}
Here, $\lambda_{\text{agn}}$ controls the influence of the agnostic mask-guided loss on the total loss.

\subsection{Clothing\&Temporal Consistency Attention}
\label{sec:temporalconsistency}

As shown in \cref{fig:vvt3.3}, existing VVT methods suffer from temporal inconsistency, leading to artifacts such as clothing flickering, unnatural fabric flow, or textures misaligned with the wearer's motion. To mitigate these issues, we aim to propose a temporal consistency mechanism. Instead of introducing computationally expensive temporal consistency modules, such as flow alignment or additional temporal modules, we propose to incorporate temporal information into the attention mechanism. This approach needs to establish inter-frame connections within a video sequence while avoiding significant computational overhead. Inspired by the role of attention mechanisms in the dual U-Net structure, we explore the use of self-attention to enhance the consistency of the target garment across frames, leading to the proposed Clothing \& Temporal Consistency Attention.

The existing self-attention mechanism in the dual U-Net is defined as:
\begin{equation}
 V_i = \operatorname{Softmax} \left( \frac{Q_i K_{p_i}^T}{\sqrt{d}} \right) V_{p_i} ,
\end{equation}
where $Q_i$, $K_{p_i}$, and $V_{p_i}$ represent the query, key, and value matrices for frame $i$, $d$ is the dimension of the key vectors. Since the same operation is applied to both $K$ and $V$, we use $x_{attn}$ to represent the unified operation on $K$ and $V$ for brevity. The computation of $x_{attn}$ is as follows:
\begin{align}
X_{attn_{p_i}} &= \operatorname{Concat}(X_{attn_{i}}, X_{attn_c}) , 
\end{align}
where $X_{attn_c}$ denotes the target garment features from ReferenceNet.

To incorporate additional temporal information into 
$X_{attn_{p_i}}$, we build upon the original cross-frame attention mechanism:
\begin{align}
X_{attn_{crossframe}} &= \operatorname{Concat}(X_{attn_{0}}, X_{attn_{i - 1}}). 
\end{align}

Through extensive observation, although significant human motion may happen, the target garment remains fixed, preventing abrupt changes, inconsistencies in the generation are primarily caused by accumulated errors. Additional temporal information is best accomplished by selecting frames with a reasonable temporal interval. Besides, unlike autoregressive methods that depend on previous frames to infer subsequent ones, our approach eliminates the need for $X_{attn_{i - 1}}$. To maintain stability, we retain $X_{attn_{i}}$.
Instead of incorporating $X_{attn_{0}}$, we introduce temporal information through a heuristic rule that selects frames based on adjacent frame differences. For frames with small $i$, the model tends to gather temporal information from later frames, aligning them with subsequent content. Conversely, for frames with large $i$, the model prioritizes earlier frames to maintain consistency with previous content. This dynamic selection strategy prioritizes temporal consistency over strict temporal order, proving more effective than fixedly selecting the $0$-th frame.

Finally, this approach effectively captures alignment information from both the current frame and temporally distant frames, even in scenarios where the garment image displays a front view while the individual enters from a side or back view. The formulation is defined as:
\begin{equation}
\begin{split}
X_{attn_{p_i}} &= \operatorname{Concat}(X_{attn_{i}}, X_{attn_{j}}, X_{attn_c}), \\
&\quad j \in \{0, \dots, i-2, i, \dots, N-1\},
\end{split}
\end{equation}
where $j$ denotes the index of a randomly selected frame and $X_{attn_c}$ represents the features extracted from the reference clothing image.

By integrating temporal references into the clothing-specific attention mechanism, our approach significantly improves temporal consistency. This ensures that the individual consistently appears to wear the same garment throughout the video sequence, regardless of variations in perspective or motion.

\begin{algorithm}[H]
\caption{Pose-guided Long VVT}
\label{alg:longvvt}
    \textbf{Input:} Agnostic video $\mathbf{A}=\{a_i\}^F_{i=1}$, Agnostic mask $\mathbf{M}=\{m_i\}^F_{i=1}$, DensePose frames $\mathbf{P}=\{p_i\}^F_{i=1}$, sample parameters $d_{pose}$, $s_{max}$\\
    \textbf{Output:} Video sequences generated $\mathbf{V}=\{V_i\}^F_{i=1}$\\
    \textbf{Step 1: Keyframe Selection and Generation} \\
\textbf{Initialize}   keyframe index $\Omega=[0]$ and $i = 0$, $j = 1$
\textbf{while} $j < F$  and \textbf{do} \\
\hspace*{1em} Compute L2 distance $\|p_i - p_j\|_2$ \\
\hspace*{1em} \textbf{if} $\|p_i - p_j\|_2 < d_{pose}$ \textbf{or} $|i - j| < s_{max}$ \textbf{then} \\
\hspace*{2em}$\Omega.$\texttt{insert}$(i).$\texttt{sort}$()$, $i = j$, \\
\hspace*{1em} \textbf{do} $j += 1$ \\
    Generate keyframes $\mathcal{V} = \{v_0, \dots, v_L\}$  \\
    \textbf{Step 2: Agnostic Keyframes Replacement} \\
\textbf{Replace} $v_i \mapsto a_i$ \textbf{if} $i \in \Omega$\\
\textbf{Divide original} $[\mathbf{A}, \mathbf{M}, \mathbf{P}]$  \textbf{into segments} $\{s_1, s_2, \dots, s_n\}$\\
    Iteratively generate and concatenate to complete $\mathbf{V}$\;
\end{algorithm}

\subsection{Pose-guided Long VVT} 
\label{sec:longvvt}
Building on enhanced spatial and temporal consistency, RealVVT generates high-quality, fixed-length video virtual try-on sequences of $N$ frames. To extend this to longer videos, we introduce a zero-shot keyframe selection strategy inspired by video translation tasks (\eg, Rerender-A-Video~\cite{yang2023rerender}). The generated keyframe outputs are interpolated as agnostic video latent features to iteratively produce the remaining try-on results, enriching the agnostic mask region with additional contextual information without directly replacing the video latent features.

\noindent\textbf{Pose-guided Keyframes Selection}.
While Rerender-A-Video~\cite{yang2023rerender} uses uniform keyframe sampling, recent methods~\cite{yang2024fresco, huang2024lvcd} sample frames based on similarity, which may not be optimal for VVT. The key factor is the magnitude of pose and motion changes, as background or facial variations can hinder accurate motion estimation. To address this, we measure object motion using the L2 distance between DensePose frames. DensePose frames have a monochromatic background (RGB: 65, 0, 82) and distinct color blocks for body parts, reducing the impact of facial or background changes. This eliminates the need for blurring techniques (\eg, Gaussian Blur) used in prior methods to suppress high-frequency texture changes. Frames with DensePose distances below a threshold $d_{dense}$ and strides under $s_{max}$ are selected as input frames.

\noindent\textbf{Agnostic Keyframes Replacement}.
We store latent features of keyframe outputs and interpolate them into the original agnostic video sequence, replacing corresponding keyframe agnostic video latents to iteratively complete the generation. For sequences with keyframes exceeding $N$, the video is divided into overlapping segments, where overlapping agnostic frames in subsequent segments are replaced with denoised results from previous segments.
\section{Experiments}
\label{sec:experiment}
\noindent\textbf{Dataset}.
We conduct the experiments using two image-based virtual try-on datasets, VITON-HD~\cite{choi2021viton} and DressCode~\cite{morelli2022dress} , and two  publicly available video datasets, ViViD~\cite{vivid_fang2024vivid}, VVT~\cite{bai2022single}. 
VITON-HD focuses on upper garment virtual try-on and provides high-resolution image pairs for garment swapping. DressCode is a comprehensive high-resolution dataset containing diverse clothing categories, including tops, lower-body garments, and dresses, for both men and women. ViViD, a recent video-image pair dataset, offers a resolution of 832 × 624. To enhance the stability of video generation, we jointly train on both image and high-resolution video datasets. For fair comparison with baselines, we train our model at a resolution of 512 × 384. we conducted evaluations on the
VVT dataset on a uniform resolution of 512 x 384,  and evaluations on VITON-HD and DressCode at a resolution of 1024 × 768.

Furthermore, for any missing inputs required by these four datasets, such as agnostic video, agnostic mask, or densepose, we generate them using Detectron2~\cite{detectron2}  and SAPIENS~\cite{khirodkar2025sapiens}.

\begin{figure}[t]
  \centering
    \includegraphics[width=0.95\linewidth]{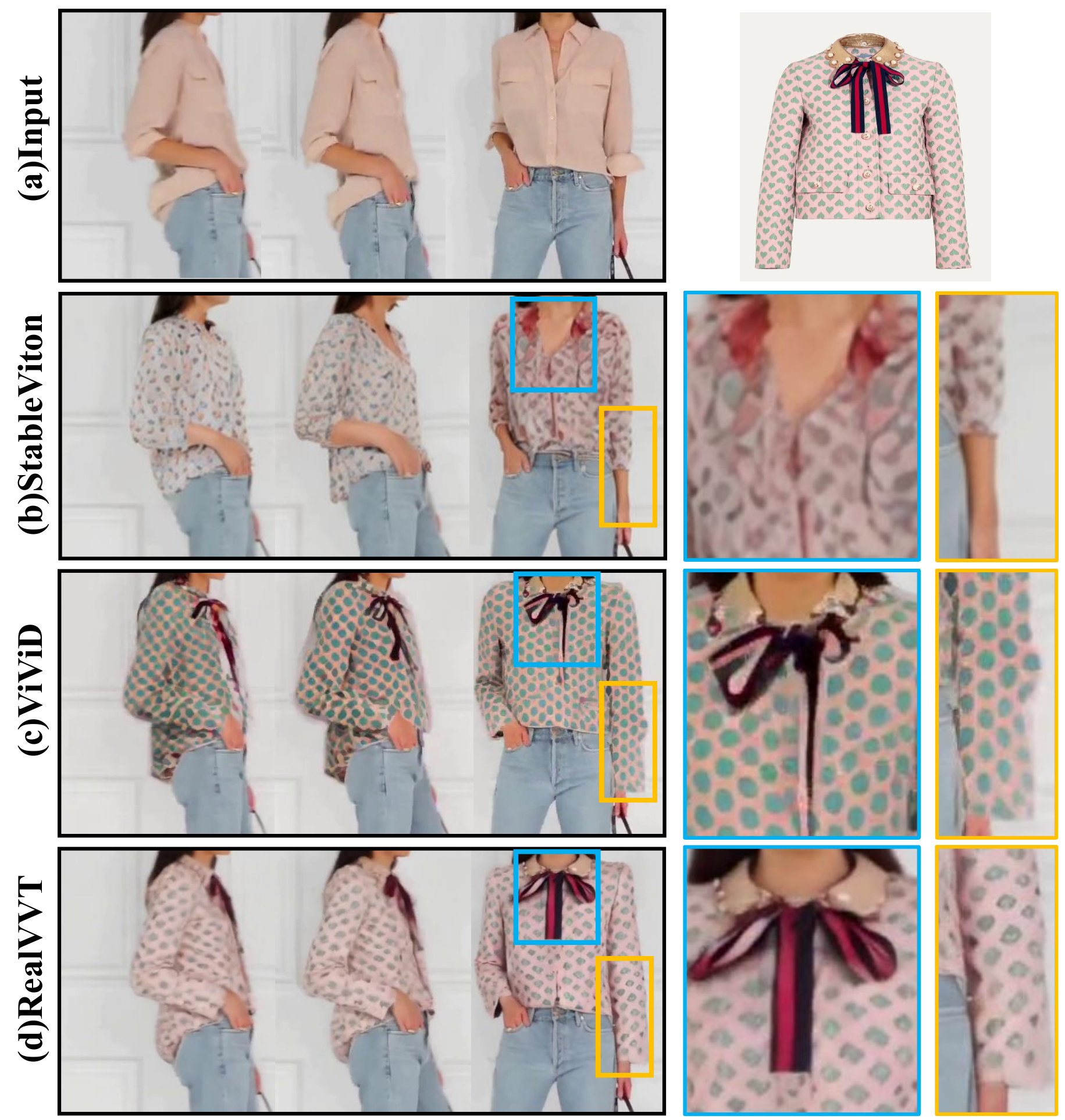}

   \caption{Virtual try-on comparison with state-of-the-art methods. (b) StableViton, an image-based VTO method, exhibits significant flickering in continuous video generation. (c) ViViD, a video-based VTO method, suffers from unstable clothing appearance, particularly noticeable around the neckline, as well as visible artifacts.  (d) Our method ensures consistent clothing appearance with realistic texture preservation and minimal artifacts.
   }
   \label{fig:vvt3.3}
   \vspace{-0.1cm}
\end{figure}

\noindent\textbf{Implementation details}.
\label{sec:implementationdetails}
The experiment is conducted on eight NVIDIA Tesla A800 GPU. We set batch size $=$ 2 based on the input video resolution, the learning rate is set to $5e^{-5}$. 
For the details in \cref{eq:loss5} and \cref{eq:loss4}, the agnostic mask loss weight,  $\lambda_{\text{agn}}$, is set to 0.5, and the scale factor, $\lambda_N$, is set to 0.01. $\lambda_N$ determines the proportion of negative samples, calculated as the sum of all background tokens (out of mask), while the positive sample is only one token corresponding to the maximum value. Given a 19×12 feature map size corresponding to a training resolution of 512×384, the number of negative tokens is, on average, approximately 120 times that of the single positive token.
Therefore, $\lambda_N$ is fixed at 0.01.The backbone we use is a combination of ReferenceNet and a Denoising UNet, pre-trained with Stable Diffusion 2.1 and Stable Video Diffusion XT respectively. 

\begin{table}[h]
\centering
\resizebox{0.9\linewidth}{!}{
\begin{tabular}{l|c|c|c|c}
\toprule
Method & SSIM $\uparrow$ & LPIPS $\downarrow$ & $\text{VFID}_{\text{I3D}}$ $\downarrow$ & $\text{VFID}_{\text{ResNeXt}}$ $\downarrow$   \\
\midrule 

StableVITON~\cite{kim2024stableviton}  & 0.876 & 0.076 & 4.021 & 5.076 \\
ClothFormer$^*$~\cite{jiang2022clothformer} & 0.921 & 0.081 & 3.97 & 5.05 \\
Tunnel Try-on$^*$~\cite{xu2024tunnel} & 0.913 & \underline{0.054} & 3.345 & 4.614 \\
VITON-Dit$^*$~\cite{zheng2024viton} & 0.896 & 0.080 & \underline{2.498} & \underline{0.187} \\
ViViD~\cite{vivid_fang2024vivid} & \underline{0.949} & 0.068 & 3.405 & 5.074 \\
WildVidFit~\cite{wild_vid_fit_he2024wildvidfit} & - & - & 4.202 & - \\
GPD-VVTO$^*$~\cite{wang2024gpd} & 0.928 & 0.056 & \textbf{1.28} & - \\
Ours & \textbf{0.976} & \textbf{0.037} & 2.689  & \textbf{0.0913}\\
\bottomrule
\end{tabular}}
\caption{Quantitative comparison on the VVT dataset. Methods marked with $^*$ are trained on additional private video data, while other methods are trained on publicly available datasets. \textbf{Bold} and
\underline{underline} denote the best and the second best result, respectively. The following tables are presented in the same way.}
 \vspace{-0.2cm}
\label{tb:quantitative_evaluation_vvt}
\end{table}

\subsection{Comparison with State-of-the-Art Methods}
\label{sec:stateoftheart}
\noindent\textbf{Metrics}. For quantitative evaluation, We evaluate our method on both a video dataset, VVT, and two image datasets, VITON-HD and Dresscode. 

We follow the  video generation evaluation paradigm of VITON-Dit~\cite{zheng2024viton} by using SSIM, LPIPS, $\text{VFID}_{\text{I3D}}$ and $\text{VFID}_{\text{ResNeXt}}$ scores. For image generation quality assessment, we use SSIM, LPIPS, FID and KID scores followed StableVITON~\cite{kim2024stableviton}.

\begin{figure}[t]
  \centering
    \includegraphics[width=1.0\linewidth]{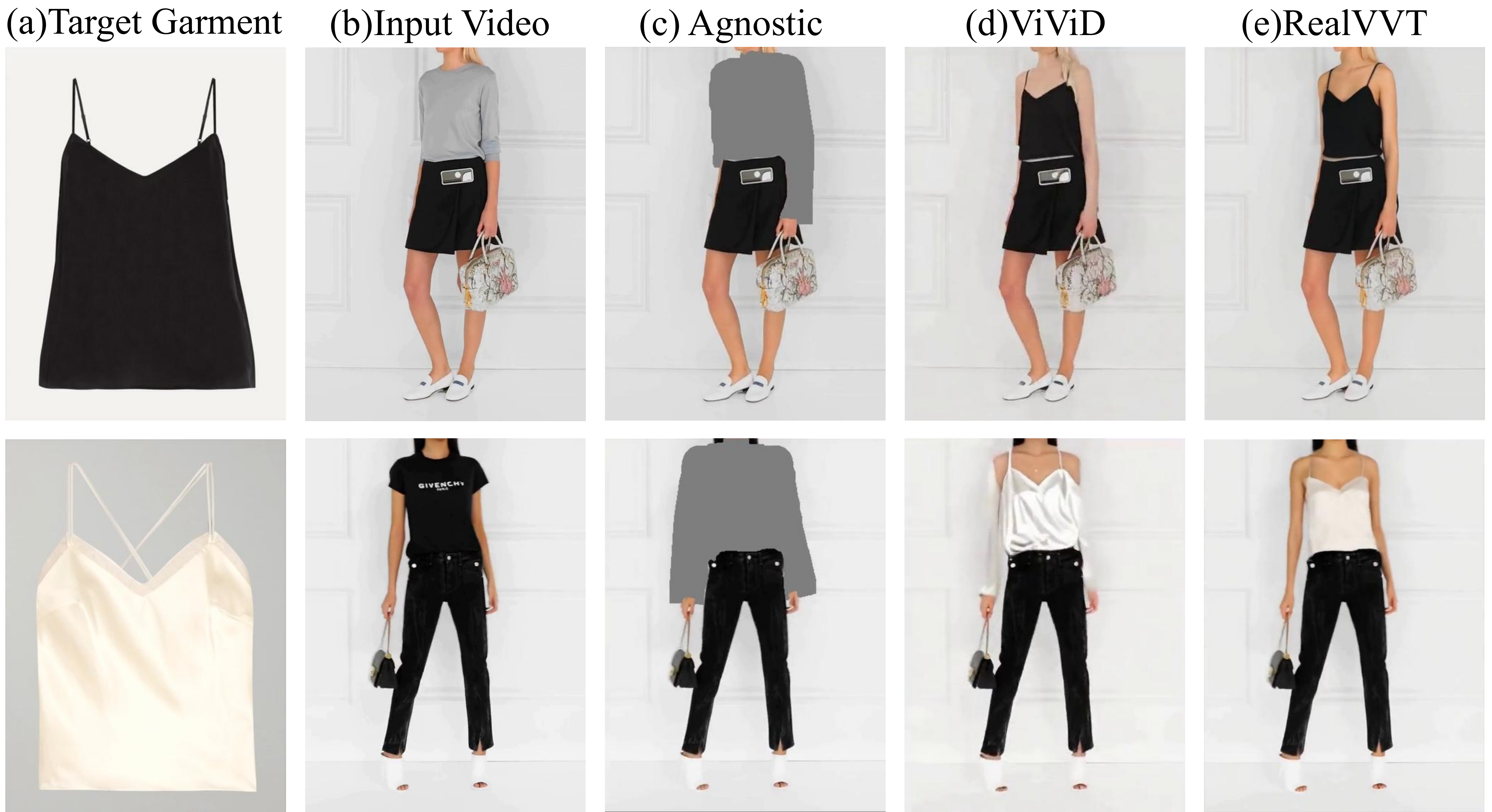}
   \caption{Virtual try-on results for a challenging case: fitting a small garment onto a large agnostic mask video. Comparisons are shown between ViViD and RealVVT, both trained at 512×384 resolution and tested at 832×624 resolution.}
   \label{fig:the4}
   \vspace{-0.2cm}
\end{figure}

\noindent\textbf{Video Dataset Evaluation}. 
Our evaluation results on the VVT dataset, as shown in \cref{tb:quantitative_evaluation_vvt}, demonstrate that our method generally outperforms prior approaches. Notably, it achieves significant improvements in SSIM, LPIPS, and $\text{VFID}_{\text{ResNeXt}}$, highlighting its effectiveness in generating spatially and temporally coherent results with preserved details.

The visualization results of the VVT dataset are omitted here due to its low resolution, which is further degraded when zooming in on clothing. Instead, we provide visual comparisons on the VIVID dataset, as shown in \cref{fig:vvt3.3} and \cref{fig:the4}. 
\cref{fig:vvt3.3} (a) displays the original video and the input target garment. This example was chosen because the target garment has longer sleeves and more complex textures compared to the original outfit. In \cref{fig:vvt3.3} (b), the results from StableVITON exhibit significant flickering. In contrast, the video-based method ViViD, shown in \cref{fig:vvt3.3} (c), greatly improves temporal coherence by eliminating flickering. However, it still fails to fully capture realistic clothing dynamics, such as the natural movement of the tie, and struggles to preserve the target garment's color, dot pattern, collar design, and tie style. In \cref{fig:vvt3.3} (d), our approach achieves superior consistency in maintaining the color, shape, and fine details of the clothing.

In \cref{fig:the4} (d), ViViD often fills the gap between the agnostic mask and the target garment with ill-suited content, such as incorrect skin tones (top) or artifacts like unnatural sleeve extensions (bottom). Our results in  \cref{fig:the4} (e) show that the Agnostic Mask-Guided loss effectively prevents such issues, ensuring the accurate generation of the garment's expected shape. 

\begin{table}[h]
\centering
\resizebox{0.7\linewidth}{!}{
\begin{tabular}{l|c|c|c|c}
\toprule
Method & SSIM $\uparrow$ & LPIPS $\downarrow$ & FID $\downarrow$ & KID $\downarrow$ \\
\midrule
CP-VTON~\cite{wang2018toward} & 0.785 & 0.2871 & 48.86 & 4.42 \\
HR-VTON~\cite{lee2022high} & 0.878 & 0.0987 & 11.80 & \underline{0.37} \\
LaDI-VTON~\cite{morelli2023ladi} & 0.871 & 0.0941 & 13.01 & 0.66 \\
DCI-VTON~\cite{gou2023taming} & 0.882 & \underline{0.0786} & 11.91 & 0.51 \\
WildVidFit~\cite{wild_vid_fit_he2024wildvidfit} & \underline{0.883}  & \textbf{ 0.0773} & \underline{8.67} & 0.51 \\
Ours & \textbf{ 0.890 } & 0.101  & \textbf{ 7.844 } & \textbf{ 0.151 } \\
\bottomrule
\end{tabular}}
\vspace{-0.1cm}
\caption{Quantitative comparison on the VITON-HD dataset.}
\label{tb:quantitative_evaluation_vitonhd}
\end{table}

\begin{table}[h]
\centering
\resizebox{0.7\linewidth}{!}{
\begin{tabular}{l|c|c|c|c}
\toprule
Method & SSIM $\uparrow$ & LPIPS $\downarrow$ & FID $\downarrow$ & KID $\downarrow$ \\
\midrule
CP-VTON~\cite{wang2018toward} & 0.820 & 0.2764 & 57.70 & 4.56 \\
HR-VTON~\cite{lee2022high} & 0.924 & 0.0605 & 13.80 & 0.28 \\
LaDI-VTON~\cite{morelli2023ladi} & 0.915 & 0.0620 & 16.71 & 0.61 \\
GC-DM~\cite{zeng2024cat} & 0.915 & 0.0649 & 14.91 & 6.01 \\
WildVidFit~\cite{wild_vid_fit_he2024wildvidfit} & \underline{0.928} & \textbf{0.0432} & \underline{12.48} & \underline{0.19} \\
GPD-VVTO~\cite{wang2024gpd} & - & - & 10.11 & 0.28 \\
Ours & \textbf{0.932} & \underline{0.0608} & \textbf{8.881}  & \textbf{ 0.163 } \\
\bottomrule
\end{tabular}}
\caption{Quantitative comparison on DressCode-Upper dataset.}
\label{tb:quantitative_evaluation_dresscode_upper}
\vspace{-0.2cm}
\end{table}

\begin{table}[h]
\centering
\resizebox{0.7\linewidth}{!}{
\begin{tabular}{l|c|c|c|c}
\toprule
Method & SSIM $\uparrow$ & LPIPS $\downarrow$ & FID $\downarrow$ & KID $\downarrow$ \\
\midrule
PBE~\cite{yang2023paint} & 0.804 & 0.2108 & 22.44 & 6.78 \\
MGD~\cite{baldrati2023multimodal} & 0.893 & 0.0689 & 13.67 & 3.79 \\
LaDI-VTON~\cite{morelli2023ladi} & \underline{0.910} & \textbf{0.0596} & 13.76 & 4.61 \\
GC-DM~\cite{zeng2024cat} & 0.902 & \underline{0.0621} & \underline{10.25} & 1.81\\
GPD-VVTO~\cite{wang2024gpd} & - & - & 11.02 & \underline{0.69} \\
Ours & \textbf{ 0.912 } &  0.0743  & \textbf{ 9.204 } & \textbf{ 0.256 }  \\ 
\bottomrule
\end{tabular}}
\caption{Quantitative comparison on DressCode-Lower dataset.}
\label{tb:quantitative_evaluation_dresscodelower}
\vspace{-0.2cm}
\end{table}

\begin{table}[h]
\centering
\resizebox{0.7\linewidth}{!}{
\begin{tabular}{l|c|c|c|c}
\toprule
Method & SSIM $\uparrow$ & LPIPS $\downarrow$ & FID $\downarrow$ & KID $\downarrow$ \\
\midrule
PBE~\cite{yang2023paint} & 0.761 & 0.2516 & 30.04 & 18.44 \\
MGD~\cite{baldrati2023multimodal} & 0.844 & 0.1195 & 12.14 & 2.41 \\
LaDI-VTON~\cite{morelli2023ladi} & 0.854 & \underline{0.1076} & 13.00 & 4.05 \\
GC-DM~\cite{zeng2024cat} & \underline{0.863} & 0.1091 & 10.71 & 2.02 \\
GPD-VVTO~\cite{wang2024gpd} & - & - & \underline{10.46} & \underline{0.70} \\
Ours & \textbf{ 0.888 } & \textbf{ 0.0932 } & \textbf{ 10.45 } & \textbf{ 0.239 } \\
\bottomrule
\end{tabular}}
\caption{Quantitative comparison on DressCode-Dresses dataset.}
\vspace{-0.2cm}
\label{tb:quantitative_evaluation_dresscode_dresses}
\end{table}

\noindent\textbf{Image Dataset Evaluation}. 
 We evaluated our model on four distinct datasets, including the VITON-HD test set and the DressCode test subsets (LowerBody, UpperBody, and Dresses). For these four datasets, we compared our method against the best-performing approaches available that provide these evaluation metrics in \cref{tb:quantitative_evaluation_vitonhd}, \cref{tb:quantitative_evaluation_dresscode_upper}, \cref{tb:quantitative_evaluation_dresscodelower}, and \cref{tb:quantitative_evaluation_dresscode_dresses}. Furthermore, our visual comparisons will be presented in the supplementary materials. Our results demonstrate a clear performance advantage across all datasets, affirming the robustness and adaptability of our approach in various virtual try-on scenarios.

\begin{table}[h]
\centering
\resizebox{0.9\linewidth}{!}{
\begin{tabular}{l|c|c|c|c|c}
\toprule
C\&T & $\lambda_{\text{agn}}$ & SSIM $\uparrow$ & LPIPS $\downarrow$ & $\text{VFID}_{\text{I3D}}$ $\downarrow$ & $\text{VFID}_{\text{ResNeXt}}$ $\downarrow$   \\
\midrule
 - & 0 & 0.890 & 0.145 & 6.119 & 0.522 \\
 - & 0.05 & 0.901 & 0.150 & 6.102 & 0.531 \\
 - & 0.1 & 0.905 & 0.102 & 5.278 & 0.235 \\
 - & 0.5 & 0.910 & 0.096 & 4.761 & 0.151 \\

+ & 0.5 & \textbf{0.976} & \textbf{0.037} & \textbf{2.689}  & \textbf{0.0913}\\
\bottomrule
\end{tabular}}
\caption{Quantitative ablation study of $\lambda_{\text{agn}}$ and C\&T.}
\vspace{-0.4cm}
\label{tb:ablation}
\end{table}

\subsection{Ablation Study}
\label{sec:ablation}

\begin{figure}[t]
  \centering
    \includegraphics[width=0.9\linewidth]{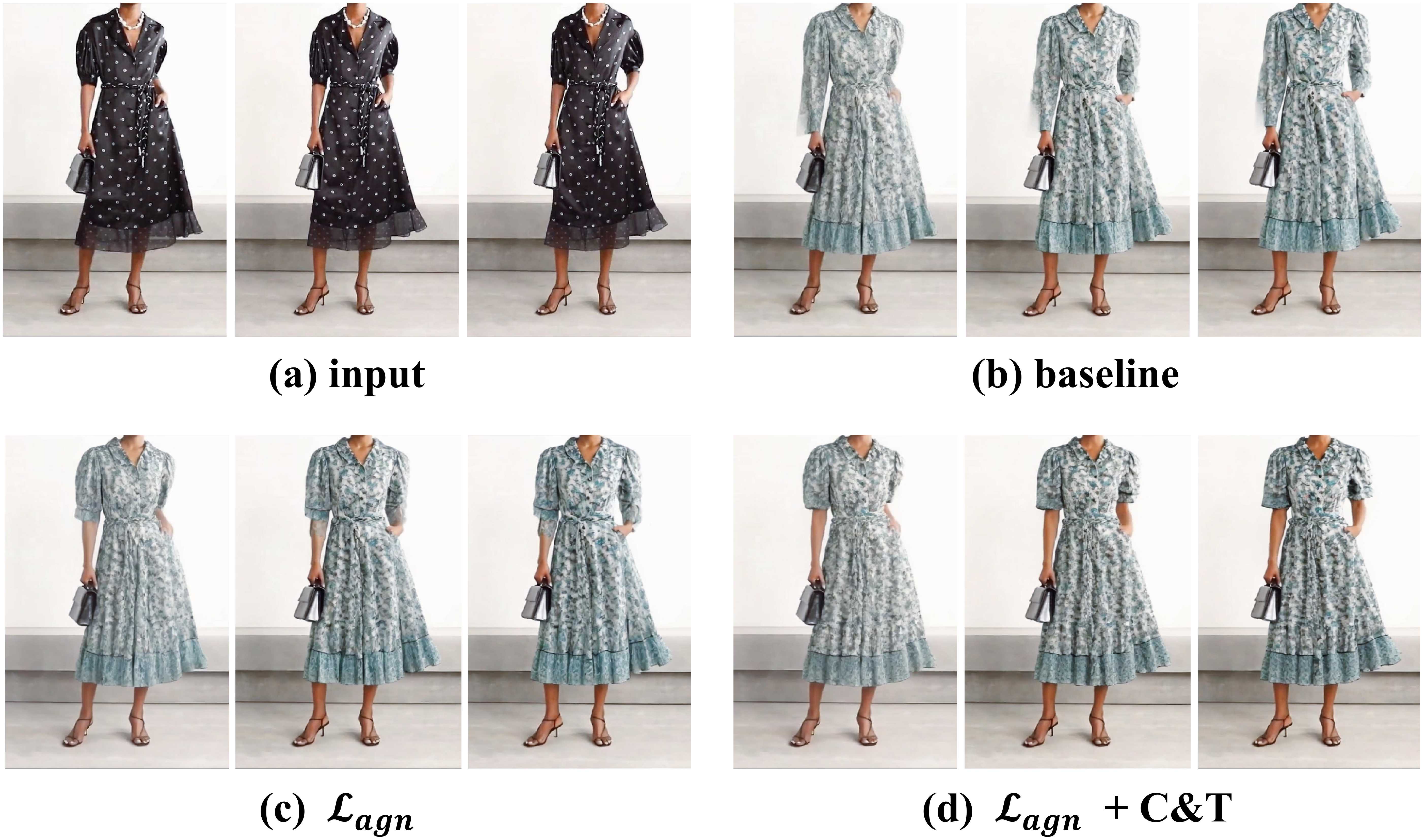}
   \caption{Effect of Agnostic Mask-Guided loss and Clothing \& Temporal Consistency Attention. The first and third images in (a) are input frames, while the second image is not used as input and instead serves to illustrate the original video. }
   \label{fig:ablation}
   \vspace{-0.65cm}
\end{figure}

To demonstrate the effectiveness of RealVVT in addressing spatial and temporal consistency, we investigate the impact of two proposed components during training: the Agnostic Mask-Guided loss ($\lambda_{\text{agn}}$), with a fixed $\lambda_N$ of 0.1, and the Clothing \& Temporal Consistency Attention mechanism (C\&T). \cref{fig:the4} show the visualization results. And \cref{tb:ablation} shows the quantitative performance improvements after adding these components. 
The first four rows in \cref{tb:ablation} present results without C\&T, with each successive row indicating a larger $\lambda_{\text{agn}}$ value. In the second row, with $\lambda_{\text{agn}} = 0.05$, the agnostic loss has minimal impact, resulting in stable metrics with no significant changes. The $\text{VFID}_{\text{I3D}}$ value shows a slight decrease, while $\text{VFID}_{\text{ResNeXt}}$ slightly increases, attributed to minor variations due to training uncertainty. As $\lambda_{\text{agn}}$ increases, both quantitative metrics improve, indicating that the model becomes more resilient to the impact of the agnostic loss, successfully generating detailed features like accurate sleeve shapes. As shown in \cref{fig:ablation} (c), compared with \cref{fig:ablation} (b), the model overcomes challenging occlusions from the large agnostic mask in the initial frames, producing the expected short-sleeved dress. However, temporal consistency across the video remains suboptimal, as the short sleeves gradually "extend" into long sleeves in subsequent frames. 

Finally, C\&T and the setting with $\lambda_{\text{agn}} =0.5$ improve  $\text{VFID}_{\text{I3D}}$ and $\text{VFID}_{\text{ResNeXt}}$ decrease significantly by 2.072 and 0.0597, respectively, indicating enhanced stability. Visual results in \cref{fig:ablation} (d) confirm improved frame-to-frame continuity and greater garment consistency across the sequence, demonstrating the effectiveness of the combined C\&T and agnostic mask-guided loss.

\section{Conclusion}
\label{sec:conclusion}

We present RealVVT, a novel framework designed to generate highly accurate virtual try-on videos with robust spatial and temporal consistency. Built on dual U-Net structure, RealVVT introduces several key innovations, including the Clothing \& Temporal Consistency Attention, Agnostic Mask-Guided Loss, and Pose-guided Long VVT. Experimental results demonstrate that RealVVT achieves state-of-the-art performance, producing high-quality virtual try-on videos with exceptional temporal coherence and garment consistency. Furthermore, the framework generates high-resolution, photorealistic images, making it highly suitable for practical virtual try-on applications. These advancements not only push the boundaries of virtual try-on technology but also hold significant potential for enhancing realism and user engagement in e-commerce platforms, paving the way for more immersive and interactive shopping experiences.

{
    \small
    \bibliographystyle{ieeenat_fullname}
    \bibliography{main}
}

\clearpage
\appendix


\clearpage
\setcounter{page}{1}
\maketitlesupplementary

\section{More Image Dataset Visual Results}
\begin{figure*}[t]
\centering
\includegraphics[width=0.95\textwidth]{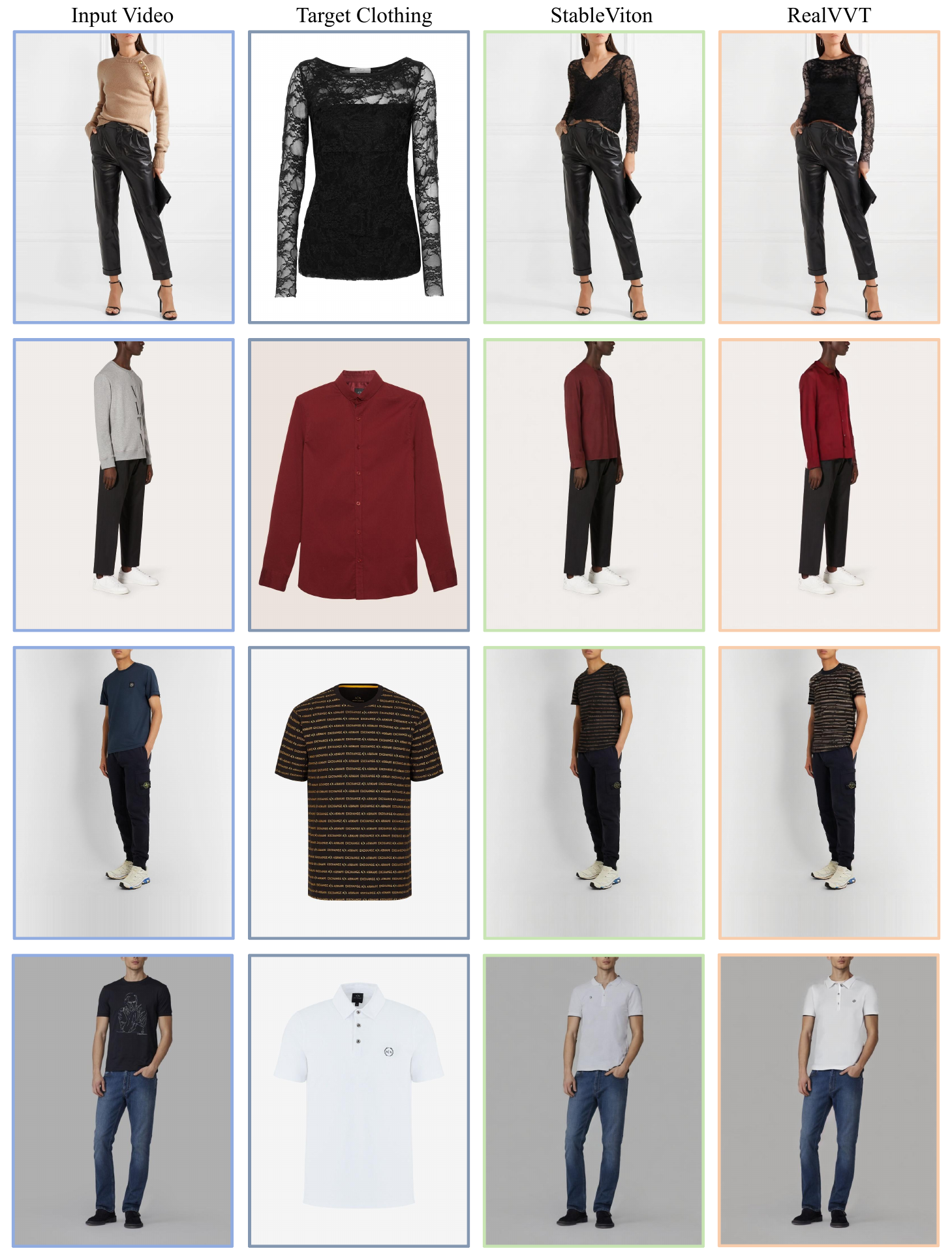}
\caption{Comparison between StableVITON~\cite{kim2024stableviton} and RealVVT on the DressCode-Upper dataset. RealVVT excels in preserving the shape and color of target garments, particularly in maintaining fine details such as collar designs.}
\label{fig:dresscodeupper}
\vspace{-0.2cm}
\end{figure*}

\begin{figure*}[t]
\centering
\includegraphics[width=0.95\textwidth]{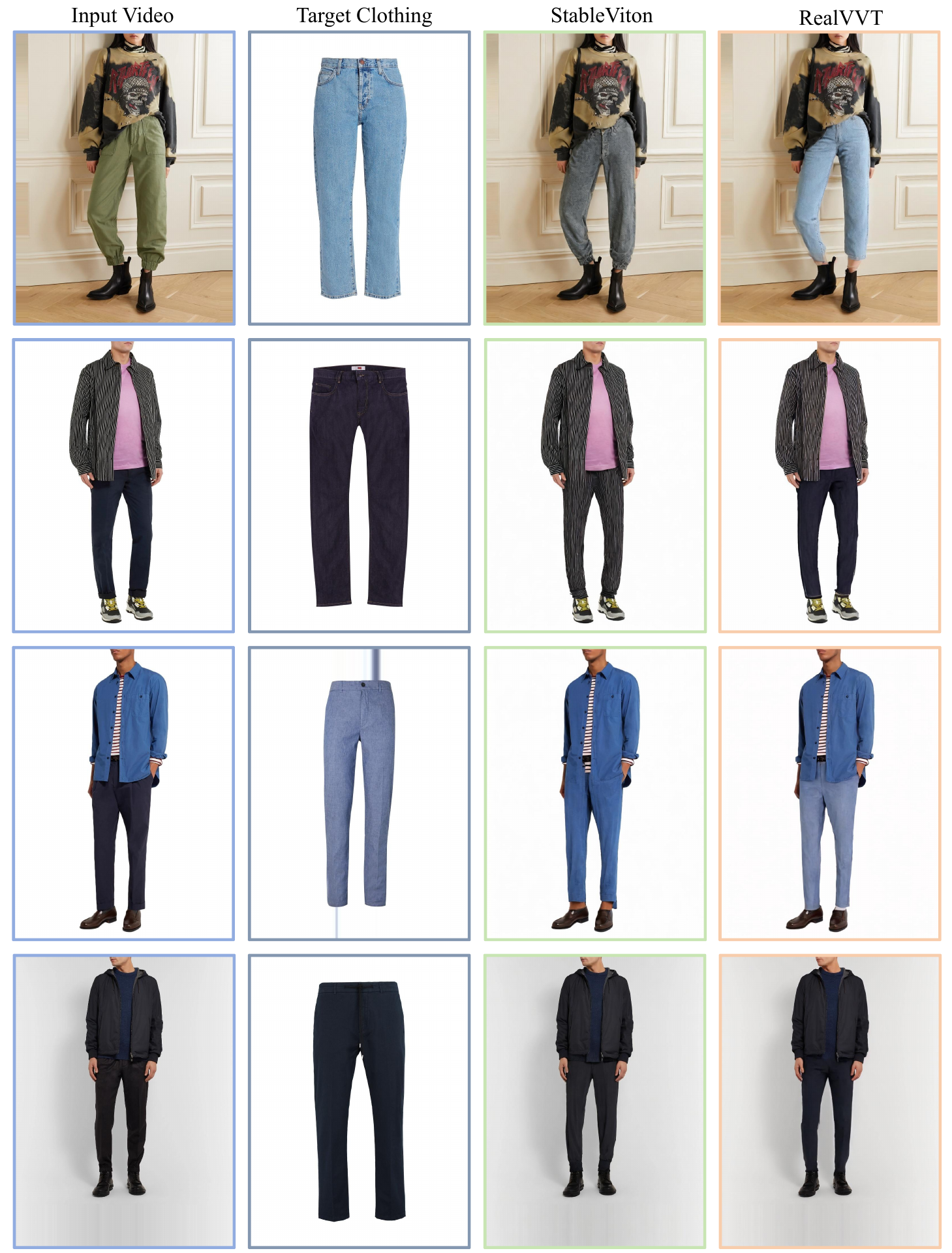}
\caption{Comparison between StableVITON~\cite{kim2024stableviton} and RealVVT on the DressCode-Lower dataset. RealVVT demonstrates superior robustness against the influence of the subject's upper body clothing and environmental factors, enabling the generated pants to seamlessly integrate into the scene while retaining their distinct characteristics.}
\label{fig:dresscodelower}
\vspace{-0.2cm}
\end{figure*}

In the experiments section, we provide quantitative comparisons with several image-based methods. In \cref{fig:dresscodeupper} and \cref{fig:dresscodelower}, we supplement these results with a visual comparison between RealVVT (ours) and StableVITON~\cite{kim2024stableviton}, a concurrent work that has gained significant recognition for addressing the virtual try-on task. Using publicly available model checkpoints, we generate try-on images for StableVITON and evaluate both methods on the DressCode dataset, the largest high-resolution image-based try-on dataset, which includes examples for upper body, lower body, and dresses, covering both female and male clothing changes.

While the main text focuses on visualizations using the ViViD dataset—a high-resolution video dataset primarily featuring female clothing changes—DressCode allows us to demonstrate RealVVT's applicability to male try-on tasks. Notably, most concurrent works, such as StableVITON, IDM-VTON~\cite{choi2024improving}, and WildVidFit~\cite{wild_vid_fit_he2024wildvidfit}, emphasize upper-body clothing during training and evaluation, likely due to the greater availability of upper-body data pairs in existing datasets (\eg, VITON-HD is exclusively an upper-body dataset, and upper-body pairs dominate DressCode compared to lower-body and dress pairs).

In \cref{fig:dresscodelower}, we showcase RealVVT's performance in handling lower-body garments. Our results highlight superior preservation of target garment color, shape, and fine details, such as trouser leg features, compared to StableVITON. To ensure fairness, we use the unpaired image pairs originally provided by DressCode without manual matching, further substantiating the robustness and generalizability of RealVVT across diverse garment types and wearer scenarios.

\section{Limitations \& Discussion}
During data collection and experimentation, we observed significant segmentation accuracy issues in existing image-based and video-based try-on datasets. These inaccuracies affect both agnostic masks and DensePose extractions, often resulting in oversized masks or masks that fail to fully cover the original clothing. Additionally, DensePose frequently suffers from incomplete limb detections, such as partially captured legs despite their visibility in the original video. Attempts to leverage state-of-the-art automated segmentation methods, such as SAPIENS~\cite{khirodkar2025sapiens} and SAM2~\cite{ravi2024sam}, yielded limited improvements and failed to provide satisfactory segmentation performance for these datasets.

Furthermore, the generation of agnostic masks and DensePose in existing video datasets often relies on image-based segmentation tools, introducing temporal inconsistencies and significant jitter across frames. While our method demonstrates robustness against such inconsistencies, considerable effort is still required to counteract the input data's inherent discontinuities and instability. Addressing these challenges underscores the need for further optimization of existing video datasets and the development of high-quality video try-on datasets, which we identify as key directions for future research.

{
    \small
}

\end{document}